\documentclass{svproc}
\usepackage{amsmath}
\usepackage{amssymb} 
\usepackage{graphicx}
\usepackage{amsfonts}
\usepackage{cite} 
\usepackage{graphics}
\usepackage{caption}
\usepackage{subfig}
\usepackage[table,xcdraw,dvipsnames]{xcolor} 
\usepackage{tabulary}
\usepackage{algorithm}
\usepackage{algpseudocode}
\usepackage{multicol}
\usepackage{lipsum}
\usepackage{mwe}
\usepackage{url}
\usepackage{wrapfig}
\usepackage{ulem}

\newcommand{\ls}[1]{{\color{Emerald}{{#1}}}} 

\begin{document}
\mainmatter              
\title{Towards Optimized Distributed Multi-Robot Printing: An Algorithmic Approach}
\titlerunning{Optimized Distributed Multi-Robot Printing: An Algorithmic Approach}  
%
\author{Kedar Karpe\inst{1} \and Avinash Sinha\inst{1} \and Shreyas Raorane\inst{1} \and Ayon Chatterjee\inst{1} \and Pranav Srinivas\inst{1} \and Lorenzo Sabattini\inst{2}}

\authorrunning{Kedar Karpe et al.} 
%
\tocauthor{Ivar Ekeland, Roger Temam, Jeffrey Dean, David Grove,
Craig Chambers, Kim B. Bruce, and Elisa Bertino}
\institute{Department of Electronics and Communication Engineering,
\\SRM Institute of Science and Technology, India\\
\email{kedarprasad\_pra@srmuniv.edu.in}
\and
Department of Sciences and Methods for Engineering (DISMI),
\\University of Modena and Reggio Emilia, Italy}

\maketitle              

\begin{abstract}
This paper presents a distributed multi-robot printing method which utilizes an optimization approach to decompose and allocate a printing task to a group of mobile robots. The motivation for this problem is to minimize the printing time of the robots by using an appropriate task decomposition algorithm. We present one such algorithm which decomposes an image into rasterized geodesic cells before allocating them to the robots for printing. In addition to this, we also present the design of a numerically controlled holonomic robot capable of spraying ink on smooth surfaces. Further, we use this robot to experimentally verify the results of this paper.

\keywords{multi-robot coordination, distributed printing}
\end{abstract}

\section{Introduction}
\label{section:intro}
Multi-robot systems have presented themselves as a systematic means of performing large and complex tasks by leveraging coordination among comparatively simpler robots. Based on the team composition, such systems are generally classified in two ways: homogenous and heterogeneous multi-robot teams. Homogenous teams comprise identical robots from both a hardware and control perspective. Whereas, heterogeneous teams can comprise of two or more types of robots. While both kinds of systems have their own set of advantages in respective applications, we are particularly interested in homogenous multi-robot systems.


In our work, we consider the problem of large-format printing using a group of homogenous mobile robots. Since distributed printing is not a widely addressed topic in the literature, we direct our study mainly to task decomposition and coverage control algorithms for multi-robot systems~\cite{kong2006distributed, sabattini2016hierarchical, gerkey2003multi, gerkey2004formal}. \cite{kong2006distributed} discusses a distributed coverage control algorithm based on Boustrophedon decomposition. The work aims at achieving coverage of virtually bounded areas for applications such as lawn mowing, chemical spill clean-up, and humanitarian de-mining. Looking at a different perspective, \cite{sabattini2016hierarchical} considers the problem of coordinating AGVs inside a warehouse. The work presents a way to partition a map into smaller sectors such that the problem of coordination between AGVs is localized only to the intersection of such sectors, thus reducing interactions between the vehicles.

In this paper, we present an optimized distributed multi-robot printing system for printing large images on smooth surfaces. Additionally, we present the design of a holonomic mobile robot to perform such printing experiments. Some of the preliminary results of our work are also presented in~\cite{karpe2019sprinter}. However, the major contribution of this paper is the task decomposition, which divides an image into geodesic cells, similar to \cite{sabattini2016hierarchical, digani2019coordination}, before allocating them to each robot. The algorithm builds upon the classical clustering problems studied vastly in the field of data science. Our work assumes the rasterization of images to be an optimal printing method over any other toolpath planning schemes and hence, the algorithm presents itself as a scalable means of decomposing images for allocating them to a group of robots.

K.H Lee and J.H. Kim have addressed a similar distributed printing problem in~\cite{lee2006multi}. They present a mobile printing system (MPS), which uses a genetic algorithm to distribute printing data to the robots. However, the algorithm presented in their paper has two major drawbacks: (1) The task allocation method results in heavily overlapping trajectories. Since this method relies on an external arbitration controller for collision avoidance, the increase in the possible number of collisions directly affects the total printing time; (2) The authors assume the print graphic to be a set of spline curves. While this assumption might significantly reduce a priori computational cost, the method is not scalable for printing intricate 2-dimensional graphics. In this paper, we try to address these drawbacks by making use of non-overlapping geodesic cells for task allocation, and by assuming graphic images to be a set of pixels rather than spline curves.

\section{Problem Definition}
\label{section:problem}
Consider the problem of coordinating $\mathcal{N}$ SPRINTER robots to print a raster image of size $u \times v$ pixels. Let $\mathrm{x}_i \in \mathbb{R}^2$ be the position of each $i-th$ robot with an associated single integrator kinematic model: $\mathrm{\dot{x}}_i = \mathrm{u}_i$, where $\mathrm{u}_i$ is the control input to the robot. For simplicity, let us consider each $i-th$ robot to be a circle with an associated radius $\mathrm{r}_i$ and a random initial position $\mathrm{x}_i(0)=(x_i(0), y_i(0))$.\\

Let $T_i$ be the time contributed by each $i-th$ robot to complete its task of printing a partition of the image. Our aim is to minimize the sum of distance travelled by each robot or alternatively the sum of their individual printing time $T_i$. Additionally, since the printing time of the process is defined by the robot that takes the maximum time, the cost function should also account for distributing the task equally to each robot. Hence, we formulate our objective function as a maximization problem:

\begin{equation}
\label{eqn:obj}
\text{maximize} \frac{\prod_{i=1}^{\mathcal{N}}T_i}{\sum_{i=1}^{\mathcal{N}}T_i} = \text{maximize} (\frac{T_1 \times T_2 \times T_3 \times ... \times T_\mathcal{N}}{T_1 + T_2 + T_3 + ... + T_\mathcal{N}})
\end{equation}

The denominator term in equation~\ref{eqn:obj} ensures that the total printing cost is minimized, whereas, the numerator term helps in distributing the task equally to every robot.\\

Since the printing algorithm considers rasterized images, for simplicity, let us assume an image to be a set of pixels $P = \{p_k: k=1, 2, 3, ..., u\times v\}$. Each pixel $p_k$ has an associated position $\mathrm{x}_k \in \mathbb{R}^2$ and an associated binary value $\delta_k$. A unity value of $\delta_k$ corresponds to an ink spray at the pixel's position, and is zero otherwise\footnote{In this paper, we consider only binary raster images due to hardware limitation of the robot. But this work can be easily extended to higher dimensional images simply by normalizing pixel data.}. Hence, we define the set of `printable' pixels as $P_1 = \{p_k: \delta_k=1, k \in P\}$ and a set of `non-printable' pixels as $P_0 = \{p_k: \delta_k = 0, k \in P\}$.

The individual cost function $T_i$ of each robot can be computed as a linear function of the the total distance traveled by the robot or alternatively, the number of pixels traversed, in both sets $P_1$ and $P_0$. We define the robot velocity as a piecewise linear function:

\begin{equation}
\mathcal{V}(p_k) = %
  \begin{cases}
      \mathcal{V}_{travel} & p_k \in P_0 \\
      \mathcal{V}_{print} & p_k \in P_1
   \end{cases}
\label{eqn:timecost}
\end{equation}

\begin{figure}
	\centering
	\includegraphics[width=0.6\columnwidth]{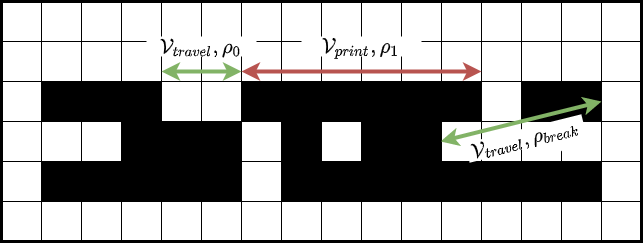}
	\caption{Velocity description of the robot at various instances of path traversal.}
	\label{fig:velo}
\end{figure}

Figure~\ref{fig:velo} depicts the robot velocities graphically. We define our individual cost function of each robot as:
\begin{equation}
\label{eqn:indcost}
T_i = \frac{\sum \rho_{1}}{\mathcal{V}_{print}} + \frac{\sum \rho_{0} + \sum \rho_{break}}{\mathcal{V}_{travel}}
\end{equation}

where, $\rho_{1}$ is the distance along `printable' pixels, $\rho_{0}$ is the distance along `non-printable pixels' and $\rho_{break}$ is the distance traveled between consecutive rows.

In the forthcoming sections, we discuss a geodesic cellularization method for clustering the pixels of an image such that the objective function in Equation~\ref{eqn:obj} is satisfied. 

\section{Geodesic Cellularization}
\label{section:cells}
As outlined in the previous sections, we partition our image into geodesic cells which are the outcome of a modified k-means clustering algorithm, before allocating them to the robots. The objective of our clustering algorithm is to allocate printing data to the robot in such a manner that requirements of the formulation in Equation~\ref{eqn:obj} are met.

Since the primary goal of this paper is to present a distributed printing system, we believe that having even a suboptimal solution for Equation~\ref{eqn:obj} is adequate. Moreover, since the pixel distribution across multiple images can be highly undeterministic, modeling a generalised clustering problem that satisfies Equation~\ref{eqn:indcost} is quite complex. Hence, we resort to using a suboptimal method, and empirically provide performance guarantees for the system.

Our geodesic cellularization approach is based on the well known k-means clustering algorithm~\cite{duda1973pattern}. By minimizing the within-cluster variances of `printable' pixels, we minimize the magnitude of quantity affected by $\mathcal{V}_{travel}$ in Equation~\ref{eqn:indcost}. In addition to the general k-means problem, we impose an additional constraint that ensures each geodesic cell has an equal number of `printable' pixels, thereby maintaining the cost function in Equation~\ref{eqn:obj}.

Since we have $\mathcal{N}$ robots available, our aim is to obtain $\mathcal{N}$ geodesic cells from a given image. Consider $\mathrm{x}_m$ as the position of $m-th$ pixel $p_m \in P_1$ of the image and $\mu_n$ as the mean of the $n-th$ cluster, and let $\mathcal{M}$ be the cardinality of the set $P_1$. Using Lemma 2.1 in~\cite{bradley1997clustering}, we reformulate the k-means clustering problem as a bilinear program:

\begin{align}
\underset{\mu, W}{\text{minimize}} \quad & \sum_{m=1}^{\mathcal{M}} \sum_{n=1}^{\mathcal{N}} W_{mn}(\frac{1}{2} \|\mathrm{x}_m - \mu_n\|^{2}_{2}) \label{eqn:cnst1}\\
\textrm{subject to} \quad & \sum_{n=1}^{\mathcal{N}} W_{mn} = 1 , m = 1, ..., \mathcal{M} \label{eqn:cnst2}\\
  & W_{mn} \geq 0, m = 1, ..., \mathcal{M}; n = 1, ..., \mathcal{N} \label{eqn:cnst3}\\
& \sum_{m=1}^{\mathcal{M}} W_{mn} \geq \left \lfloor \frac{\mathcal{M}}{\mathcal{N}} \right \rfloor , n = 1, ..., \mathcal{N} \label{eqn:cnst4}
\end{align}

In Equation~\ref{eqn:cnst1}, $W_{mn}$ is the selection variable whose binary value represents if the $m-th$ pixel belongs to the $n-th$ cell. The constraint in Equation~\ref{eqn:cnst4} imposes bounds on the number of pixels in a cell thereby equalizing the number of 'printable' pixels in each geodesic cell.

We solve the formulation in Equation~\ref{eqn:cnst1} using an iterative refinement technique given by Lloyd's algorithm. Since the solution of the k-means clustering problem is known to be highly sensitive to its initial conditions, we initialize the cluster means using k-means++ seeding algorithm~\cite{arthur2006k}. Thus, the solution of Equation~\ref{eqn:cnst1} can be obtained iteratively in 3 steps:

\begin{enumerate}
\item \textbf{Cluster Initialization:} Initialize the cluster means $\mu_n(t=0)$, where t is the number of iterations, using k-means++ algorithm.\\

\item \textbf{Cluster Assignment:} With $W_{mn}(t)$ as the solution to the linear programming problem at the $t-th$ iteration, evaluate Equation~\ref{eqn:cnst1}.\\

\item \textbf{Cluster Update:} Update cluster means $\mu_n (t+1)$:
\begin{equation}
\mu_n (t+1)=  %
  \begin{cases}
     \frac{\sum_{m=1}^{\mathcal{M}}W_{mn}(t)\mathrm{x}_m}{\sum_{m=1}^{\mathcal{M}}W_{mn}(t)} & if \sum_{m=1}^{\mathcal{M}}W_{mn}(t)>0\\
\mu_n (t) & otherwise
   \end{cases}
\end{equation}
\end{enumerate}

\begin{figure}
	\centering
	\includegraphics[width=\columnwidth]{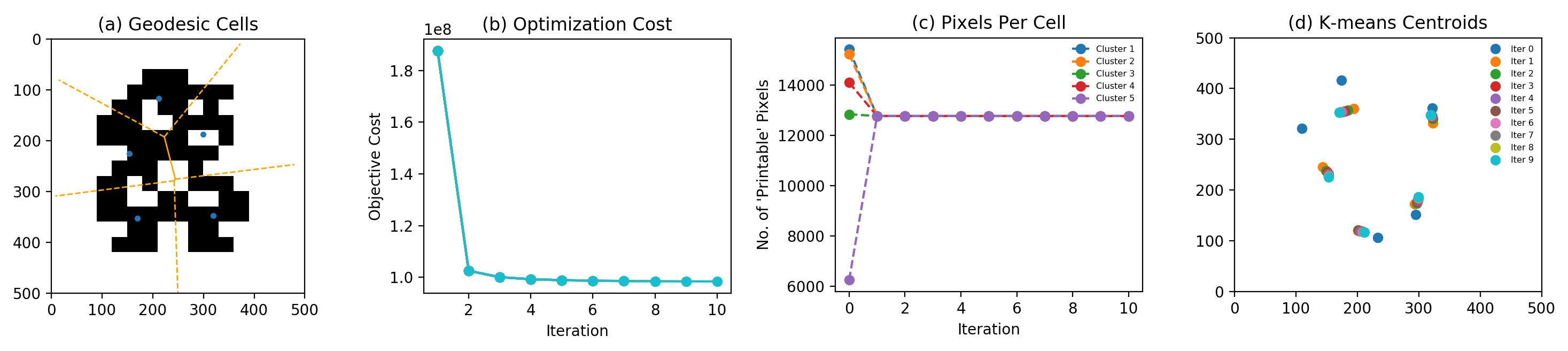}
	\caption{This figure represents the results of an geodesic cellularization experiment which uses constrained k-means clustering.}
	\label{fig:pymip}
\end{figure}

After the initialization in Step 1, iterate over Step 2 and Step 3 until the cluster means stabilize. Figure~\ref{fig:pymip} shows results from a clustering experiment on a sample image. The solution of the experiment is shown as geodesic cells and their respective means in Figure~\ref{fig:pymip}(a). Figure~\ref{fig:pymip}(b) shows the stabilization in the cost function for up to 10 iterations. Figure~\ref{fig:pymip}(c) represents the number of `printable' pixels in each cell while Figure~\ref{fig:pymip}(d) shows the cluster means over the 10 iterations.

In Section~\ref{section:sims}, we present detailed empirical proofs which suggest this method to be a suboptimal solution for the formulation in Equation~\ref{eqn:obj}.

\section{Cell Assignment}
\label{section:assign}
The goal of performing cell assignment is primarily to obtain a collision-free path between the robots' initial positions and their respective cells, whilst minimizing the sum of the travel costs of the robot. Additionally, we also ensure that these paths are collision-free. Such assignment problems have been generally solved in the literature as a linear assignment problem of minimizing the sum of distances traveled by robots~\cite{smith2007target}. These linear problems can be easily solved in polynomial time, but they do not guarantee collision-free paths. \cite{turpin2013trajectory} presents a variation of the centralized assignment problem which minimizes the sum of integral of squared velocities instead of the sum of distances, and additionally guarantees collision-free paths if a bound on the initial positions of the robots is satisfied.

In this work, we consider minimizing the sum of integral of robots' squared velocities, which is similar to minimizing the sum of squared distances between the initial and goal positions, for two main reasons:
\begin{enumerate}
\item Minimizing the sum of squared velocities ensures collision-free paths as presented in~\cite{turpin2013trajectory}.
\item Additionally, since this cost function is strictly convex, it leads to a lower sum of squared distances compared to the linear assignment problem, thus reducing the perturbation caused to the printing cost.
\end{enumerate}

Even though this method leads to a quadratic assignment problem which is NP-hard, suboptimal algorithms can be used to obtain a solution in polynomial time for a small number of robots.

The solution to the allocation problem seeks an $\mathcal{N} \times \mathcal{N}$ assignment matrix $\phi$ such that:
\begin{equation}
\label{eq:phi}
\phi_{i,j} \in \{0,1\}, \,\,\,\, \forall \,\, i,j \in \{1,2,3,...,\mathcal{N}\}
\end{equation}
We define an extended assignment matrix $\Phi = \phi \otimes I_\ls{2} $, where $I_n$ is an identity matrix in $n$ dimensions and $\otimes$ is the Kronecker product. Let $\mathrm{x}_i(t)$ be the position of the $i-th$ robot at time t, and $\mu_j$ be the position of the $j-th$ goal, which is also the centroid of the $j-th$ geodesic cell. We define the stacked position vectors as $X(t) = {[{\mathrm{x}_1(t)}^\intercal,{\mathrm{x}_2(t)}^\intercal,...,{\mathrm{x}_\mathcal{N}(t)}^\intercal}]^ \intercal$ and $C = {[{\mu_1}^ \intercal,{\mu_2}^ \intercal,...,{\mu_\mathcal{N}}^ \intercal]}^ \intercal$, where $X(t) \in  \mathbb{R}^{2\mathcal{N}}$ and $C \in \mathbb{R}^{2\mathcal{N}}$ respectively. Our aim is to find $\mathcal{N}$ finite time trajectories for the robots $\gamma(t):[t_0,t_f] \to X(t)$, where, $t_0$ and $t_f$ are the initial and final times respectively. We define additional boundary conditions on the initial and final positions of the robots and also ensure each robot is mapped only to one cell:

\begin{equation}
\label{eq:boundary}
\begin{split}
\gamma(t_0)&=X(t_0) \\
\Phi \gamma(t_f) &= C
\end{split}
\end{equation}

\begin{equation}
\label{eq:constraint}
\phi^\intercal \phi = I_{\mathcal{N}}
\end{equation}

In the solution to the assignment problem given by~\cite{turpin2013trajectory}, clearance requirements for the trajectories are ignored. Rather, the authors set a bound on the separation between the initial positions which guarantee that the generated trajectories are always collision free. Under the assumption that clearance requirements do not exist, we formulate the assignment problem as:
\begin{equation}
\begin{split}
\underset{\phi,\gamma(t)}{\text{minimize}} \int_{t_0}^{t_f} \dot{X}(t)^\intercal \dot{X}(t)dt 
\\subject ~to \,\,\,\, (\ref{eq:phi}),(\ref{eq:boundary}), (\ref{eq:constraint})
\end{split}
\label{eq:integral}
\end{equation}

Equation~\ref{eq:integral} can be reformulated as a linear assignment problem in $\Phi$ and solved using Hungarian algorithm~\cite{kuhn1955hungarian}. We then obtain the trajectories for the robots using:
\begin{align}
\gamma(t)&=
\left(  
1-\frac{t-t_0}{t-t_f}
\right)
X(t_0) +
\left(  
\frac{t-t_0}{t-t_f}
\right)
(\Phi C+(I_{Nn}-\Phi\Phi^\intercal)X(t_0))
\end{align}

\subsection{Clearance Requirements for Collision Free Trajectories}

Given $r_i$ as the radius of the $i-th$ robot, collision free trajectories can be guaranteed if:
\begin{equation}
\begin{aligned}
\|\mathrm{x}_i(t_0) - \mathrm{x}_j(t_0)\| > 2\sqrt{2}r_i && where, i,j \in \{1,2,3,...,\mathcal{N}\}
\end{aligned}
\end{equation}

For an analytical proof of freedom from collision, refer to Lemma 1 in~\cite{turpin2013trajectory}.

%
%
%

\section{Simulations}
\label{section:sims}

In this section, we present the outcomes of simulations of the proposed multi-robot printing method. 

\subsection{Distributed Printing by Evolutionary Algorithm}
With the objective of comparing the evolutionary distributed printing algorithm in~\cite{lee2006multi} to the algorithm presented in this paper, we performed simulations on a fixed set of sample images. To keep the comparison fair, we make the following assumptions:

\begin{enumerate}
\item Both methods use an equal number of robots.
\item The robots in both methods have the same printing resolution.
\item Velocities of the sets of robots in both cases are identical.
\end{enumerate}

\begin{figure}
  \centering
  \begin{tabular}[b]{c}
    \includegraphics[width=.3\linewidth]{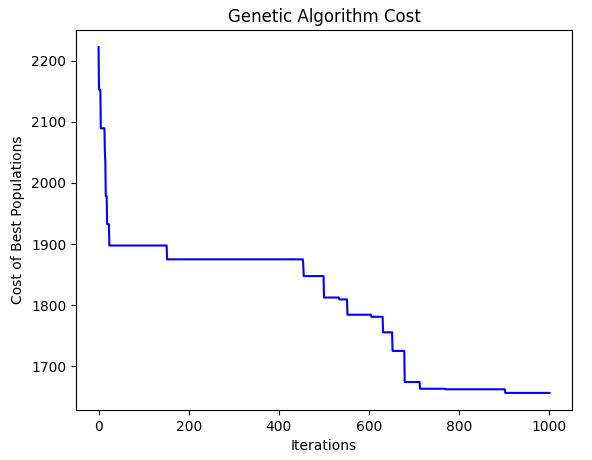} \\
    \small (a)
  \end{tabular}
  \begin{tabular}[b]{c}
    \includegraphics[width=.3\linewidth]{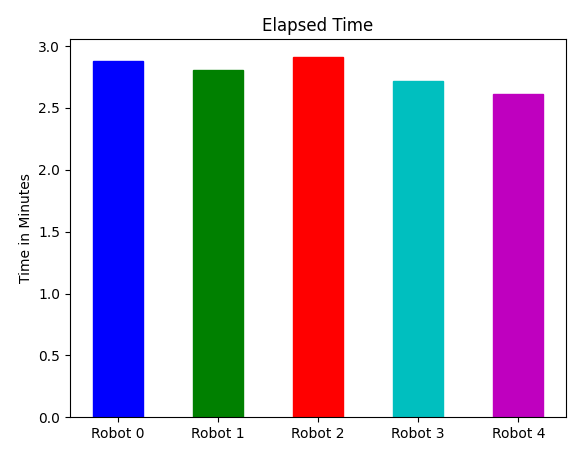} \\
    \small (b)
  \end{tabular}
\\
  \begin{tabular}[b]{c}
    \includegraphics[width=.3\linewidth]{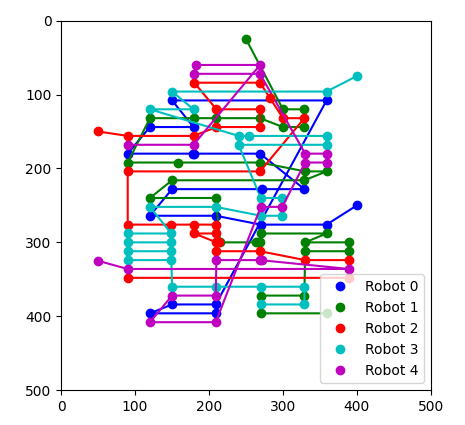} \\
    \small (c)
  \end{tabular}
  \begin{tabular}[b]{c}
    \includegraphics[width=.3\linewidth]{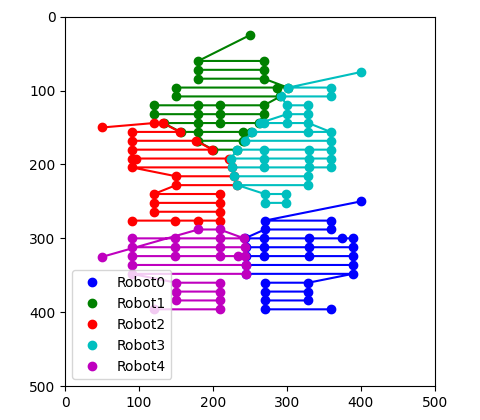} \\
    \small (d)
  \end{tabular}
  \caption{(a) Cost function of the genetic algorithm, (b) Individual printing cost of each robot in the simulation,  (c) Robot trajectories of distributed printing using evolutionary algorithm, (d) Robot trajectories of distributed printing using geodesic cellularization.}
\label{fig:lee}
\end{figure}

Figure~\ref{fig:lee}(a) shows the cost of the genetic algorithm converges to its minima with consecutive iterations. The trajectories of all the robots for the entire duration of the printing process are shown in Figure~\ref{fig:lee}(c). Note how the trajectories overlap, thus increasing the probability of collision arbitrations.

\subsection{Distributed Printing by Geodesic Clustering}
\begin{figure}[h]
  \centering
  \begin{tabular}[b]{c}
    \includegraphics[width=.33\linewidth]{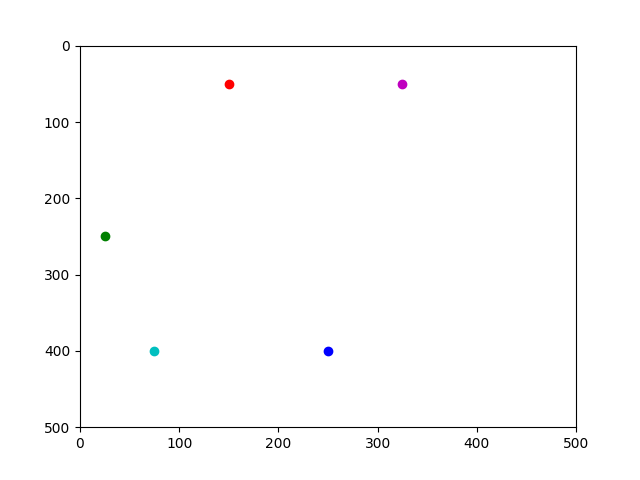} \\
    \small (a) Progress: 0\%
  \end{tabular}
  \hspace{-10px}
  \begin{tabular}[b]{c}
    \includegraphics[width=.33\linewidth]{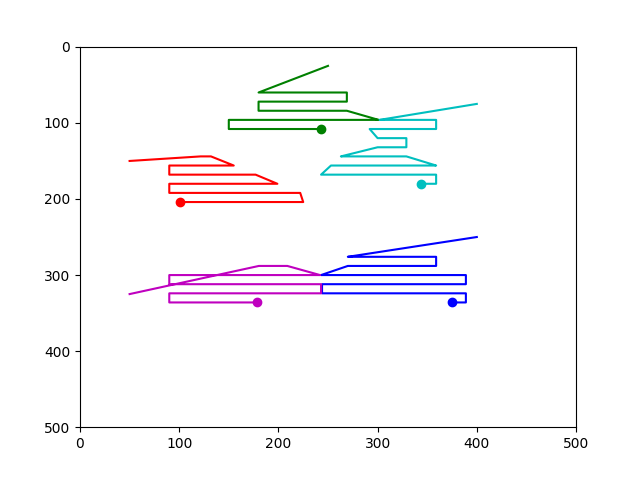} \\
    \small (b) Progress: 50\%
  \end{tabular}
  \hspace{-10px}
  \begin{tabular}[b]{c}
    \includegraphics[width=.33\linewidth]{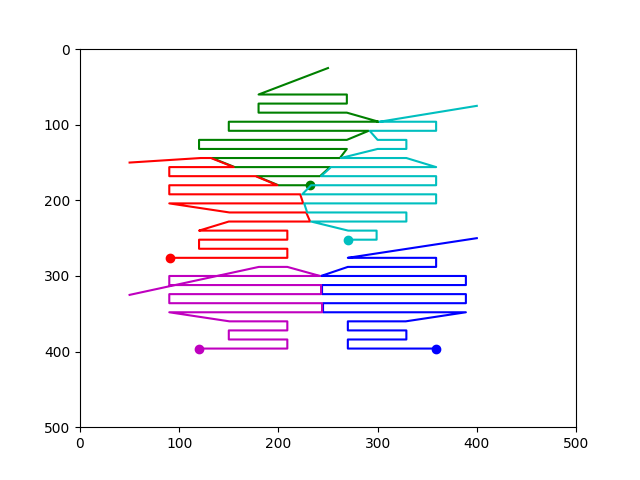} \\
    \small (c) Progress: 100\%
  \end{tabular}
  \caption{Trajectories of robots at various durations of the simulation.}
  \label{fig:prog}
\end{figure}

Using the results of the cellularization algorithm in Section~\ref{section:cells}, we developed visual simulations to verify the distributed printing method presented in this paper. Each robot rasterizes its respective cell and moves in discrete steps using the Bresenham's Algorithm. The simulation stops when all the robots have reached the last row of the rasterized image. The printing progress at different time intervals is shown in Figure~\ref{fig:prog}.

Figure~\ref{fig:lee}(d) represents a set of robot trajectories using geodesic clustering for the same image sample used in simulating the evolutionary algorithm. For the comparison between the two methods to be unbiased, we used identical simulation parameters to generate the trajectories. The primary takeaway from Figure~\ref{fig:lee}(d) is that the collision arbitrations occur only at the boundary of each cell. Thus, they do not contribute to the total printing costs as much as the overlapping trajectories in Figure~\ref{fig:lee}(b).


\subsection{Solution to the Optimization Problem}
Since finding a solution to the optimization problem in Equation~\ref{eqn:obj} by using the properties in Equation~\ref{eqn:timecost} is quite complex, thus we leverage the clustering algorithm in Section~\ref{section:cells} to compute a suboptimal solution to the optimization problem. In this section, we empirically prove that the geodesic cellularization algorithm is indeed an affordable suboptimal solution to the formulation in Equation~\ref{eqn:obj} with respect to distributed multi-robot printing.

\begin{figure}
	\centering
	\includegraphics[width= 0.9\columnwidth]{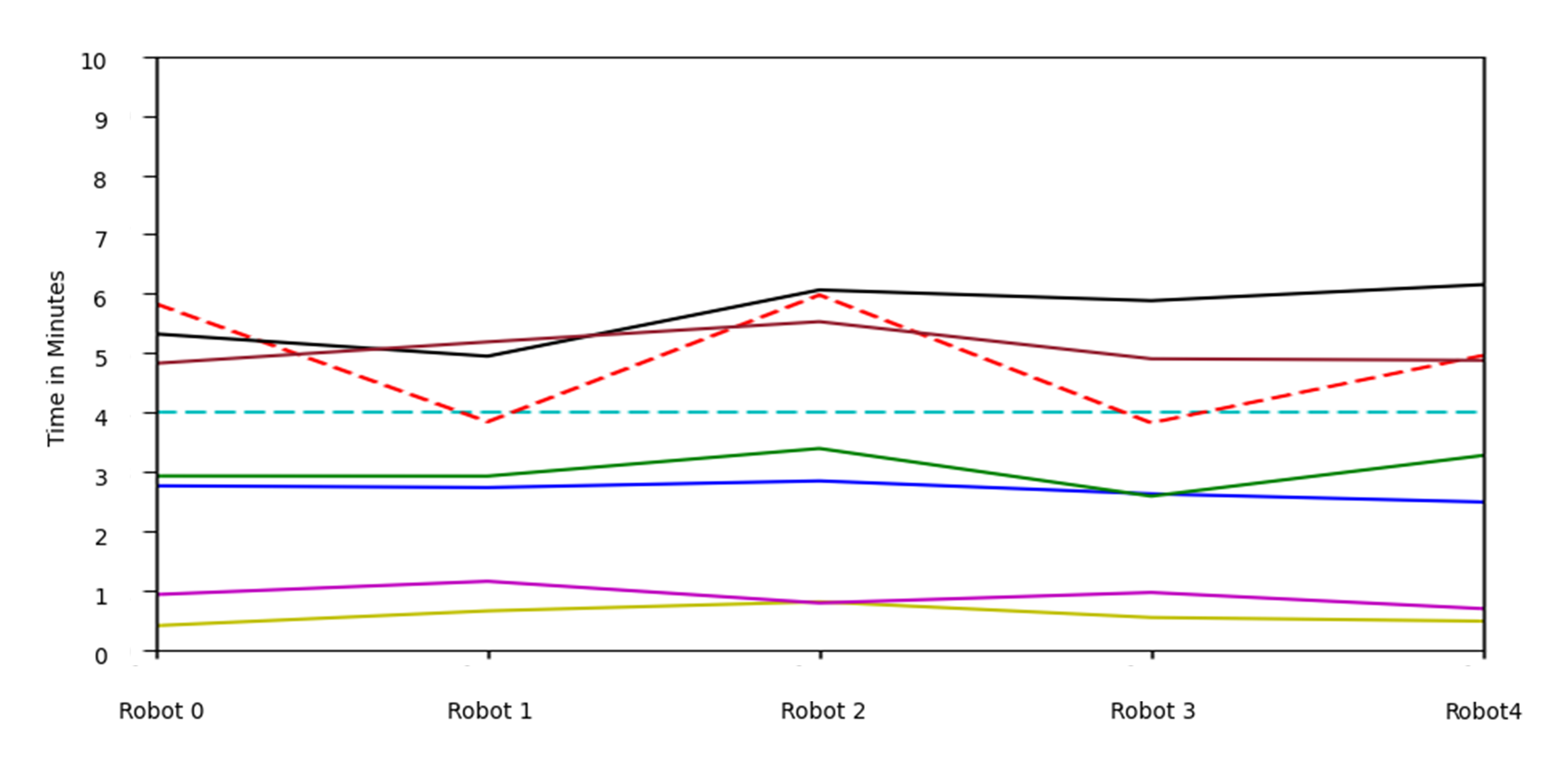}
	\caption{Printing time of each of the 5 robots for 8 distinct printing experiments.}
	\label{fig:obj}
\end{figure}

Since the total rasterized length over an image is a linear sum of the lengths of each robot's trajectories, intuitively we can say that the objective function in Equation~\ref{eqn:obj} will be maximized when the printing times taken by all robots are equal. In Figure~\ref{fig:obj}, we have presented the printing times of 5 robots for 8 distinct image samples. We can say that the optimality of these solutions obtained from the geodesic cellularization algorithm can be deduced by the `flatness' of the printing cost curves. That is, if the empirical printing times of all robots are equal, then the solution is considered optimal.

Additionally, the best case and worst case solutions are also presented in the figure. The best-case scenario would be the case where the symmetrical rotation order of the image equals the number of robots used for printing. Whereas, the worst-case scenario would be that of an image with a checkerboard pattern. We can note that even the worst-case solution presents itself to be an feasible solution for multi-robot distributed printing.

\section{Robot Design}
\label{section:robot}

We built the SPRINTER robot to perform experiments on the distributed printing algorithm presented in this paper. The robot's design brings together inkjet printing and discrete linear actuation capabilities in order to perform the printing experiments. The robot is designed to be a 4-wheeled quasi-holonomic\footnote{The platform is a 4-wheel holonomic robot, but rotation around the axis passing through the centre of the robot is restricted through software.} platform which accepts numerical control commands to deposit ink in a 2-dimensional cartesian plane.

\subsection{Mechanical Design}
\begin{figure}[h]
	\centering
	\includegraphics[width=\columnwidth]{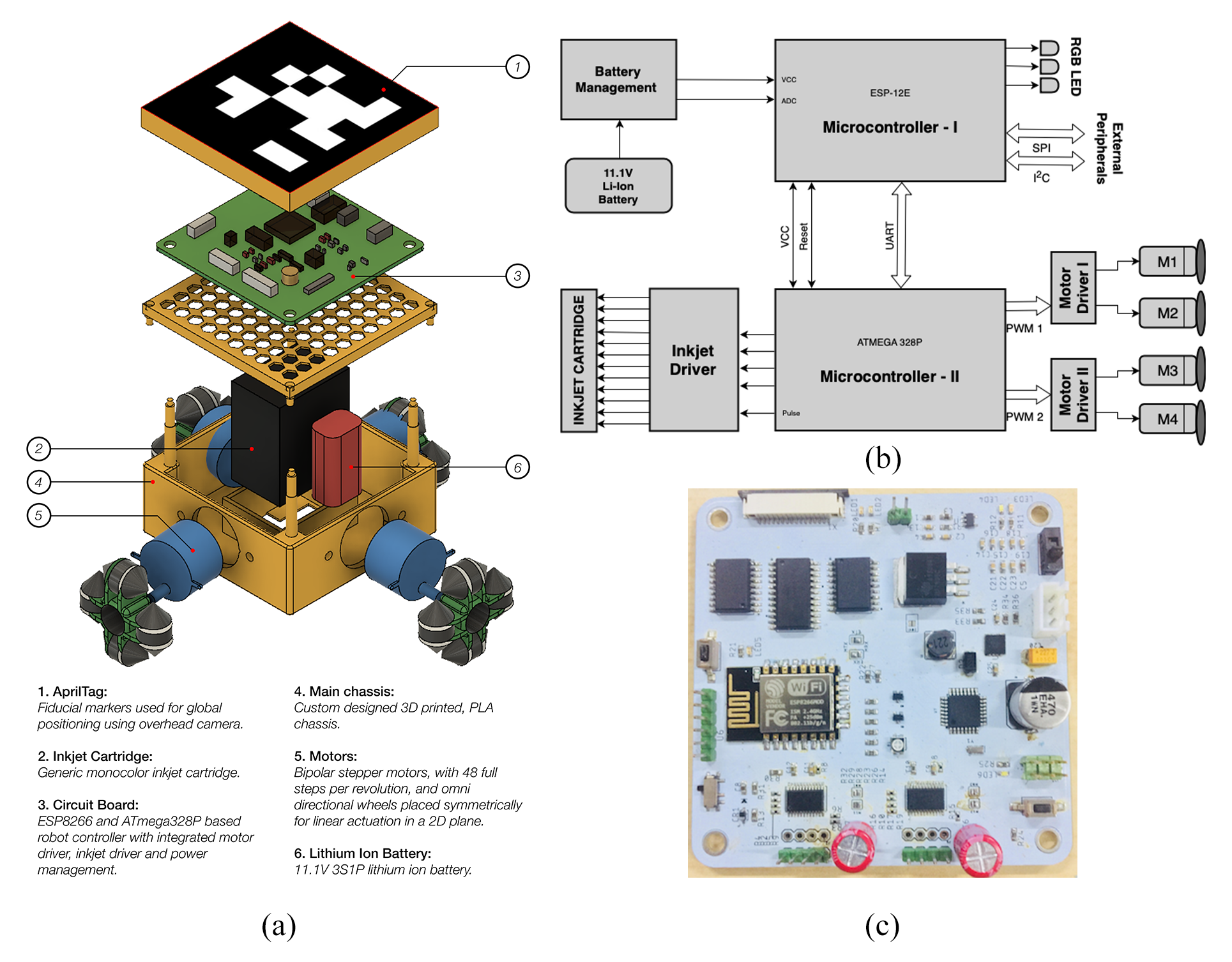}
	\caption{(a) Exploded view of the SPRINTER robot designed for printing experiments. (b) Electronics architecture of robot. (c) Custom controller circuit board of the SPRINTER robot.}
	\label{fig:hwd}
\end{figure}

The motivation behind the design of the SPRINTER robot was to experimentally verify the algorithms presented in this paper. Hence, most of the robot parts are constructed out of 3D printed materials (ABS+PLA). The robot's chassis comprises three distinct layers: a top layer, an intermediate, and a bottom layer respectively. The design of the chassis is modular, and more layers can be easily added to accommodate any additional components. Figure~\ref{fig:hwd}(a) shows an exploded view of the SPRINTER robot's design.

The top layer houses the AprilTag~\cite{olson2011apriltag} fiducial marker for global positioning. The intermediate layer holds the controller ciruit board of the robot. And the bottom layer houses the four stepper motors and a lithium-ion battery which powers the robot. The inkjet cartridge is also placed on this layer.

\subsection{Electronics Architecture}
We developed a custom controller circuit for the SPRINTER robot to accommodate all the electronics in its small footprint. The circuit was fabricated on a 2-layer printed circuit board. It mainly consists of a primary microcontroller, an auxiliary microcontroller, a motor driver circuit, and an inkjet driver. The architecture of the controller board is shown in Figure~\ref{fig:hwd}(b). All the electronics draw current from an on-board lithium polymer battery which also drives the inkjet cartridge and the stepper motors.

\subsection{Motion Discretization}
\begin{wrapfigure}{R}{0.5\textwidth}
    \begin{minipage}{0.5\textwidth}
      \begin{algorithm}[H]
        \caption{Modified Bresenham’s Algorithm for integral coordinates}
        \begin{algorithmic}
          \State Given $(x_1,y_1)$ and $(x_2, y_2)$ 
		\State $\Delta{x} = x_2 - x_1 $ and $\Delta{y}=y_2 - y_1 $
		\State $j = y_1$
		\State $\bar{\epsilon} = \Delta{x} - \Delta{y}$
		\For{$i = x_1 \, to \, x_2 -1$}
		\\ \hskip\algorithmicindent Go to $(i,j)$
		\If {$\bar{\epsilon } \geq 0$} 
		\\ \hskip\algorithmicindent $j+=1$ 
		\\ \hskip\algorithmicindent $\bar{\epsilon}-=\Delta x$
		\EndIf
		\\ \hskip\algorithmicindent $\bar{\epsilon}+=\Delta y$
		\EndFor
        \end{algorithmic}
      \end{algorithm}
    \end{minipage}
\end{wrapfigure}

We use a modified version of Bresenham's Algorithm~\cite{bresenham1965algorithm} to discretize the locomotion of the robot. Bresenham's Algorithm defines a method to obtain nearly continuous trajectories using discrete motion. The modified version of this algorithm assumes the start and end coordinates to be integral multiples of the step size of the motor. Consider an initial point $(x_1,y_1)$ and destination point $(x_2,y_2)$ in the coordinate space represented as a grid having cell dimensions equal to that of the step size and the variables $\Delta x = x_2-x_1$, and $\Delta y = y_2-y_1$. We define the driving axis $DA$ as the axis being tracked and passive axis $PA$ as the axis which evolves automatically. $x$-axis is chosen as the driving axis if $|\Delta x| \geq |\Delta y|$, and $y$-axis if $|\Delta x| < |\Delta y|$. Next, we define error $\epsilon$ as the negative distance, from any point on the line joining initial and goal locations to the top edge of the grid cell at that point and $\bar\epsilon = \Delta PA \epsilon$. The Bresenham's Algorithm keeps a track of $\epsilon$ and increments along the passive axis as $\epsilon$ becomes greater than zero. This way, we're able to achieve a discrete positional control of the SPRINTER robot.

\section{Conclusion}
In this paper, we propose a task decomposition strategy for distributed printing using a group of mobile robots. We have considered image decomposition as a clustering problem and presented an algorithm to divide an image to geodesic cells of constituent pixels. Moving further, we have discussed a method to allocate these cells to the mobile robot team. Another key takeaway of this paper is the physical design and the control strategy of the SPRINTER robot which was used for the distributed printing experiments.

We also present several simulations results to validate the theoretical assertions made in this paper. Using some of these results, we contrast our distributed printing approach with another multi-robot printing system that uses evolutionary techniques to distribute the printing task among the robots.

While we believe the results presented in this paper are a propitious step towards replacing traditional large-format printing systems with more robust distributed printers, there is definitely a scope for extending this work in the future.

Although the geodesic cellularization method is a priori step for printing, this method is very computationally heavy. In the future, we would like to consider a decentralized method, which can be solved in realtime, on-board the robots.

Additionally, we would also like to accommodate collision avoidance into the cost function since the present cost formulation is sensitive to collisions caused by the type of image and the distribution of its constituent pixels.
\label{section:conc}

%
%
\bibliographystyle{spmpsci} 
\bibliography{biblio}
\end{document}